# Lexical Prompt Compression for Large Language Models:

## A Training-Free, Deterministic Pipeline with Empirical Pareto Analysis Across Eleven Task Categories

*Author: Shamin Chokshi*

*Affiliation: AI Engineer/Data Scientist at Bright Horizons / Master's Alumni at Northeastern University*

*Contact: email: shaminchokshi2000@gmail.com / chokshi.sh@northeastern.edu / phone: 8573287776*

## Abstract

Recent advances in large language models (LLMs) have made prompts increasingly large and complex. Techniques such as chain-of-thought reasoning (Wei et al., 2022) and in-context learning (Brown et al., 2020) frequently push real-world prompts past several thousand tokens, increasing inference cost and latency. Learned compression methods such as LLMLingua (Jiang et al., 2023) and Selective Context (Li et al., 2023) achieve high compression ratios but require auxiliary language models and are non-deterministic. We ask a complementary question: how far can a training-free, fully deterministic, CPU-only pipeline based on classical lexical NLP be pushed before output quality degrades significantly? Eleven toggleable lexical transformations - stopword removal, filler-phrase deletion, contraction and abbreviation substitution, part-of-speech-based pruning, lemmatization, WordNet-driven synonym shortening, and named-entity preservation - are assembled into a configurable pipeline. Fifteen configurations are evaluated on 1,242 English-only prompts from six sources (Dolly-15k, LMSYS-Chat-1M, WildChat-1M, MMLU, GSM8K, HellaSwag), spanning eleven automatically derived task categories, yielding 18,630 paired GPT-4o-mini completions. Output preservation is measured using BLEU, ROUGE-1/2/L, BERTScore-F1, and SentenceBERT cosine similarity. The most aggressive configuration achieves a mean token reduction of 40.3% (sigma = 9.2) at a BERTScore-F1 of 0.876 against the original-prompt output; a stopword-only configuration achieves 29.6% reduction at 0.913. The compression-versus-fidelity Pareto frontier is characterized per task category, with commonsense reasoning a systematic failure mode under aggressive compression. All code, prompts, and per-cell results are released for reproducibility.

## 1. Introduction

Prompts submitted to large language models have become increasingly lengthy. The integration of chain-of-thought prompting (Wei et al., 2022), few-shot in-context learning (Brown et al., 2020), retrieval-augmented contexts, and agentic system prompts routinely generates inputs comprising thousands of tokens. As most commercial LLM APIs charge per input token and inference latency increases with sequence length, this trend directly elevates costs and slows downstream applications.

The recent literature has responded with a family of learned prompt compression techniques. LLMLingua (Jiang et al., 2023) uses a small auxiliary language model (typically a GPT-2 or LLaMA-7B variant) to estimate per-token informativeness and prune low-information tokens, reporting up to 20x compression on long-context tasks. Selective Context (Li et al., 2023) takes a similar self-information approach at the lexical-unit level. LongLLMLingua (Jiang et al., 2024) extends the framework to multi-document long-context scenarios. These methods set the state of the art for raw compression ratio at fixed quality.

.

However, learned methods entail significant deployment costs: they require execution of a non-trivial auxiliary model alongside each inference, introduce non-determinism due to sampling, temperature, and seed dependence, and are typically GPU-bound. For practitioners seeking to reduce prompt length by 30–40% without incurring additional computational costs, no published baseline systematically characterizes the capabilities of a purely lexical, deterministic, CPU-only pipeline. The most closely related work consists of prompt-engineering practices such as stopword removal and abbreviation, which have not been systematically evaluated at scale.

## 1.1 Research questions

1. RQ1. How much of a prompt can be removed by classical lexical transformations alone, and how does that compression interact with output preservation?
2. RQ2. Which individual lexical techniques contribute most to compression, and which add little marginal value?
3. RQ3. Are the compression-vs-fidelity tradeoffs uniform across task categories, or are some categories systematically more or less robust to aggressive compression?
4. RQ4. Which combination of toggles yields the highest token reduction subject to a fixed semantic-similarity floor on the resulting model output?

## 1.2 Contributions

- A modular eleven-technique lexical compression pipeline implemented with NLTK (Loper & Bird, 2002), WordNet (Miller, 1995), and tiktoken, designed for full reproducibility (single random seed, no neural compression component).

- A large-scale empirical study comprising 1,242 prompts x 15 toggle configurations = 18,630 paired GPT-4o-mini completions, with zero failed cells.
- Per-task-category Pareto frontiers across sixteen categories drawn from six source datasets, with explicit identification of failure modes (notably commonsense reasoning).
- A reproducible open-source toolkit (FastAPI service, Streamlit UI, SQLite results store, batch runner, categorization pipeline, plotting suite).

We make no claim to outperform learned methods such as LLMLingua. The contribution is positional: a clean, transparent baseline against which both practitioners and future learned compressors can be compared, plus a controlled empirical map of where lexical compression succeeds and where it breaks.

# 2. Related Work

## 2.1 Prompt compression

Selective Context (Li et al., 2023) established an early formalization of prompt compression via token pruning, utilizing per-token self-information derived from a compact base language model. This methodology was further developed by LLMLingua (Jiang et al., 2023), which introduced a coarse-to-fine budget controller and iterative token-level compression to achieve significant reduction ratios across benchmarks such as GSM8K, BBH, and ShareGPT. Subsequent iterations, including LongLLMLingua (Jiang et al., 2024) and LLMLingua-2, have addressed long-context Retrieval-Augmented Generation (RAG) and task-agnostic BERT-based classification, respectively. However, these extant approaches are characterized by non-deterministic implementations and necessitate substantial computational resources, typically requiring GPU inference or intensive CPU overhead. In contrast, the present work introduces a deterministic, training-free, and CPU-optimized framework designed for immediate interpretability. Within an experimental scope confined to short- and medium-length prompts (60–2000 characters), this approach offers a distinct alternative to resource-intensive auxiliary model-based compression.

## 2.2 Lexical NLP foundations

The lexical techniques we evaluate are well-established components of pre-neural NLP pipelines. Stopword lists, part-of-speech tagging, lemmatization, and named-entity recognition are standard NLTK operations (Loper & Bird, 2002). WordNet (Miller, 1995) supplies the synonym graph for token-shortening substitution. These components have been studied for decades in information retrieval and document summarisation, but to our knowledge no published work assembles them as a configurable pipeline and measures their joint effect on LLM output fidelity at the scale reported here.

## 2.3 Evaluation metrics

We adopt the metric stack most commonly used in the compression-and-summarisation literature. BLEU (Papineni et al., 2002) measures modified n-gram precision and is included for compatibility with prior work, though it correlates only loosely with human judgment on paraphrased outputs. ROUGE-1, ROUGE-2 and ROUGE-L (Lin, 2004) capture unigram, bigram and longest-common-subsequence overlap respectively. BERTScore-F1 (Zhang et al., 2020) computes token-level greedy cosine similarity in a BERT embedding space and correlates more strongly with human judgment than n-gram metrics. We additionally report sentence-level cosine similarity over MiniLM SentenceBERT embeddings (Reimers & Gurevych, 2019) for both prompt-level and output-level comparisons.

### 2.4 Source datasets

Our evaluation set draws from six published datasets. Databricks-Dolly-15k (Conover et al., 2023) is a 15,000-record open-license instruction-following corpus authored by Databricks employees, with eight pre-assigned task categories (closed_qa, open_qa, classification, brainstorming, summarization, information_extraction, creative_writing, general_qa). LMSYS-Chat-1M (Zheng et al., 2024) is a corpus of one million real-world conversations with 25 LLMs collected from Chatbot Arena. WildChat-1M (Zhao et al., 2024) is a one-million-conversation corpus collected by Allen AI through free GPT-3.5 / GPT-4 access. MMLU (Hendrycks et al., 2021) provides 57-subject multiple-choice questions. GSM8K (Cobbe et al., 2021) provides linguistically diverse grade-school math word problems. HellaSwag (Zellers et al., 2019) tests commonsense sentence completion.

## 3. The Compression Pipeline

### 3.1 Design principles

The pipeline is built around three principles. First, every transformation is a pure function over text, so transformations can be toggled independently and composed in any order. Second, every transformation is deterministic given a fixed configuration, so re-running the pipeline on the same input yields byte-identical output. Third, no neural model is involved in compression itself (embedding models and BERT-based metrics are used only during evaluation). All transformations run in well under 100 ms per prompt on a single CPU core.

### 3.2 The eleven techniques

We instantiate eleven independently toggleable techniques. They are applied in a fixed canonical order (multi-word transforms before single-word transforms, structural cleanups last) so that downstream stages do not undo upstream changes.

| # | Technique | What it does |
|---|---|---|

| 1 | Filler-phrase removal | Regex deletion of hedges and politeness padding: 'I was wondering if you could', 'could you please', 'if it is not too much trouble', etc. |
|---|---|---|
| 2 | Abbreviation substitution | Replaces multi-word equivalents with short forms: 'for example' to 'e.g.', 'and so on' to 'etc.', 'as soon as possible' to 'asap'. |
| 3 | Contraction substitution | 'do not' to don't, 'cannot' to can't, 'I am' to I'm, etc. |
| 4 | Filler-word removal | Single-word fillers: 'basically', 'actually', 'really', 'just', 'kind', 'sort', 'somewhat'. |
| 5 | Stopword removal | NLTK English stopword list minus negations ('not', 'no', 'never') and wh-words. Named-entity spans are protected (see #10). |
| 6 | Function-word pruning | Removes articles ('a',' an',' the') and auxiliary verbs ('is', 'are', 'was', 'be', 'have', 'do', and modals). |
| 7 | POS-keep ('caveman') | Aggressive variant: keep only nouns, verbs, adjectives, numerals, wh-words, negations. All other parts of speech are dropped. |
| 8 | Lemmatization | WordNet-aware lemmatization using POS tags ('running' to 'run', 'studies' to 'study'). |
| 9 | Synonym shortening | For each content word with at least five characters, look up WordNet synonyms; if a synonym tokenizes to fewer cl100k_base tokens, substitute. |
| 10 | Named-entity preservation | Detect NEs with NLTK ne_chunk; protect their spans from stopword and function-word removal. |
| 11 | Whitespace and punctuation normalization | Collapse multiple spaces, deduplicate redundant punctuation ('!!!' to '!'), normalize long dashes. |

Two cross-cutting safety mechanisms protect against meaning-flipping transformations. (a) The stopword list explicitly excludes all common negations and wh-words; removing 'not' or 'why' would change meaning catastrophically. (b) Wh-word and trailing '?' presence are re-asserted at the end of the pipeline if the original prompt was a question, so question form survives even very aggressive compression.

### *3.2.1 Detailed description of each technique*

**Technique 1**, filler-phrase removal, targets the multi-word politeness and hedging constructions that are pervasive in conversational prompts but carry no task-relevant information. A curated list of regular expressions matches constructions such as 'I was wondering if you could', 'would it be possible to', and 'if it is not too much trouble', deleting them entirely. Because these phrases are matched as whole units with word boundaries, the technique cannot accidentally fragment a legitimate sentence; it either matches the full hedge or leaves the text untouched. This is the single most effective technique on chat-style prompts, where such hedging is most common.

**Technique 2**, abbreviation substitution, replaces verbose multi-word expressions with their conventional short forms: 'for example' becomes 'e.g.', 'that is' becomes 'i.e.', and 'as soon as possible' becomes 'asap'. The substitution table is hand-curated to include only abbreviations that are unambiguous in general English, so that meaning is preserved. The token savings per substitution are modest, but the technique is essentially lossless, making it safe to enable in almost all configurations.

**Technique 3**, contraction substitution, collapses two-word forms into standard English contractions: 'do not' becomes "don't", 'cannot' becomes "can't", 'I am' becomes "I'm". Crucially, contraction preserves negation: 'do not' and "don't" are semantically identical, so this technique is safe even on rules and constraints where negation is load-bearing. The saving is typically one token per contraction.

**Technique 4**, filler-word removal, deletes single-word discourse markers and intensifiers that add emphasis but not content: 'basically', 'actually', 'really', 'just', 'simply', 'quite'. These words are drawn from a fixed list and removed by word-boundary regex. The technique is conservative by construction because the list contains only words whose removal does not change propositional meaning.

**Technique 5**, stopword removal, is the workhorse of the pipeline. It removes high-frequency function words using the standard NLTK English stopword list (Loper & Bird, 2002), a resource rooted in the long tradition of stopword filtering in information retrieval. Two safeguards modify the raw list: all negations ('not', 'no', 'never', and similar) and all wh-words are removed from the stopword set so that they are never dropped, and named-entity spans are protected (technique 10). Stopword removal alone accounts for the majority of the token reduction achievable by the pipeline, because function words are both frequent and individually low-information.

**Technique 6**, function-word pruning, is a more aggressive cousin of stopword removal that specifically targets articles ('a', 'an', 'the') and auxiliary and modal verbs ('is', 'are', 'was', 'be', 'have', 'do', 'will', 'would', 'could', 'should', and others). Where stopword removal uses a broad pre-built list, function-word pruning is a focused grammatical filter. It overlaps substantially with stopword removal, which is why enabling both together adds little beyond enabling either alone.

**Technique 7**, POS-keep (informally 'caveman' mode), is the most aggressive transformation. Using NLTK part-of-speech tagging (Loper & Bird, 2002), it discards every token except nouns, verbs, adjectives, numerals, wh-words, and negations, reducing the prompt to a content-word skeleton reminiscent of telegraphic speech. The name reflects the resulting terse style. This technique alone produces the highest single-technique compression, but because it relies on POS-tagger accuracy, words mis-tagged as content (such as gerunds like 'wondering') occasionally survive when a human would remove them.

**Technique 8**, lemmatization, reduces inflected words to their dictionary base form using the WordNet lemmatizer (Miller, 1995), informed by the POS tag of each token so that 'running' as a verb maps to 'run' while preserving nouns correctly. Lemmatization rarely saves tokens on its own under the cl100k_base tokenizer, because base and inflected forms often tokenize to the same number of tokens; its main value is normalizing surface form before other transforms.

**Technique 9**, synonym shortening, replaces long content words with shorter near-synonyms. For each content word of at least five characters, the technique queries WordNet (Miller, 1995) for synonyms and, if a synonym tokenizes to fewer cl100k_base tokens than the original, substitutes it. As the results section documents, this technique is the least reliable in the suite: WordNet's nearest synonym is frequently semantically narrower or register-shifted, so the technique is retained primarily so that its weakness can be measured rather than recommended for production use.

**Technique 10**, named-entity preservation, is a protective modifier rather than a reductive transform. Using NLTK's named-entity chunker (Loper & Bird, 2002), it identifies spans corresponding to people, places, and organizations and exempts those tokens from stopword and function-word removal. Without this safeguard, an entity such as 'The Hague' or 'University of the Pacific' could lose its article and become ambiguous; with it, entity spans survive intact.

**Technique 11**, whitespace and punctuation normalization, is a final structural cleanup that collapses runs of whitespace, removes spaces before punctuation, deduplicates repeated punctuation ('!!!' becomes '!'), and normalizes long dashes. It is enabled by default in every configuration because it is strictly lossless and tidies the output of the more aggressive upstream transforms.

### *3.2.2 Implementation*

The techniques fall into two implementation families. The first family is rule-based and uses no external linguistic resource: filler-phrase removal, abbreviation substitution, contraction substitution, filler-word removal, and whitespace and punctuation normalization (techniques 1 through 4 and 11). Each of these is implemented with Python regular expressions operating over hand-curated lists that are hard-coded in the source. Specifically, filler phrases are stored as a list of regular-expression patterns; abbreviations and contractions are stored as explicit mapping dictionaries from the long form to the short form; and filler words are stored as a fixed set. Substitution is performed with word-boundary-anchored regular expressions so that a pattern only matches a complete token or phrase and cannot corrupt a larger word that happens to contain it. These lists were assembled by hand rather than learned, which is what makes this family fully transparent and auditable: every transformation a user sees can be traced to a specific entry in a specific list.

The second family is library-based and relies on the Natural Language Toolkit, NLTK (Loper & Bird, 2002), for its linguistic analysis: stopword removal, function-word pruning, POS-keep, lemmatization, synonym shortening, and named-entity preservation (techniques 5 through 10).

Stopword removal uses NLTK's built-in English stopword corpus, from which we programmatically subtract our protected negation and wh-word sets at load time rather than editing the corpus itself. Function-word pruning uses a small hard-coded set of articles and auxiliaries, but depends on NLTK tokenization to apply it. POS-keep and lemmatization both use NLTK's averaged-perceptron part-of-speech tagger; lemmatization additionally uses the WordNet lemmatizer (Miller, 1995), with each token's POS tag mapped to the corresponding WordNet part of speech so that the lemmatizer behaves correctly. Synonym shortening queries the WordNet lexical database (Miller, 1995) for candidate synonyms of each long content word. Named-entity preservation uses NLTK's maximum-entropy named-entity chunker to mark protected spans. In short, the second family does not reimplement any linguistic algorithm: it composes well-established NLTK and WordNet components and adds the safety logic that protects negations, wh-words, and entities.

Token counts, used both for the synonym-shortening decision and for all compression metrics, are computed with the tiktoken library under the cl100k_base encoding, which matches the tokenization used by the GPT-4 family. The required NLTK and WordNet data files are downloaded automatically on first use, so that an end user need not perform any manual setup step. Every technique is implemented as a pure function that takes text and returns text, which is what allows the techniques to be toggled independently and composed in the fixed canonical order described above. No technique uses randomness, a trained model, or a network call at compression time, which is the basis for the pipeline's determinism.

## 3.3 Toggle configurations evaluated

From the eleven techniques, we derive fifteen named configurations. Eleven of these isolate a single technique (each combined with whitespace normalization) to enable per-technique ablation. Four are stacked configurations of increasing aggression: light (filler removal + contractions), medium (light + stopwords + abbreviations), aggressive (medium + function words + lemmatization), and maximum (everything on, including POS-keep). A no-op baseline_none configuration is included as a control.

## 3.4 Implementation

The pipeline is implemented in Python 3 using NLTK 3.8 (Loper & Bird, 2002) for tokenization, POS tagging, lemmatization, and named-entity recognition; tiktoken with the cl100k_base encoding for token counting (matching GPT-4 family tokenization); the openai client library for evaluation calls; sentence-transformers (Reimers & Gurevych, 2019) for cosine similarity; and bert_score for BERTScore-F1. All experimental results presented here use OpenAI gpt-4o-mini at temperature = 0 for the evaluation calls, removing sampling variance as a confound. The pipeline, FastAPI service, Streamlit UI, batch evaluator, and SQLite persistence layer are released as a single repository.

# 4. Experimental Setup

## 4.1 Datasets

Our combined evaluation set comprises 1,242 unique English-language prompts drawn from six sources. The breakdown is as follows:

| Source | Prompts | Role |
|---|---|---|
| Dolly-15k (Conover et al., 2023) | 664 | Pre-categorized instruction following; spans seven task types |
| LMSYS-Chat-1M (Zheng et al., 2024) | 185 | Real Chatbot-Arena user prompts; auto-categorized by our taxonomy |
| WildChat-1M (Zhao et al., 2024) | 93 | Real ChatGPT user prompts; auto-categorized |
| MMLU (Hendrycks et al., 2021) | 100 | 57-subject multiple-choice factual QA |
| GSM8K (Cobbe et al., 2021) | 100 | Grade-school math word problems |
| HellaSwag (Zellers et al., 2019) | 100 | Commonsense sentence completion |
| Total | 1,242 | |

All prompts are filtered to English by a two-stage process: trust the dataset's own language field where available (LMSYS, WildChat), then verify with langdetect to catch mislabeled or unlabeled rows. Non-Latin scripts are hard-rejected before language detection. Prompts shorter than 60 characters or longer than 2000 characters are excluded as edge cases.

### *4.1.1 Description of each source dataset*

**Databricks-Dolly-15k** (Conover et al., 2023) is a 15,000 record open source instruction-following dataset authored by Databricks employees, and released under a permissive license to enable commercial use. Each record is labeled with one of eight human-defined task categories: open QA, closed QA, classification, brainstorming, summarization, information extraction, creative writing, and general QA. Dolly is shipped with these human gold standard labels, which are the backbone of our category analysis and the only source whose categories are not inferred by our own heuristic classifier. It adds 664 prompts, the largest single contribution to our evaluation set.

**LMSYS-Chat-1M** (Zheng et al., 2024) is a corpus of one million real conversations gathered from Chatbot Arena, a public platform where users interact with and compare 25 different large language models. The prompts are written by real users with real goals, so they have the kind of phrasing you'd expect from a real user with a real goal – the hedging, verbosity, and informality that lexical compression tries to address. From this we extract 185 first-turn English prompts. LMSYS prompts are especially helpful for testing compression on real-world text instead of hand-written instructions.

**WildChat-1M** (Zhao et al., 2024) is a second large corpus of authentic user conversations, comprising one million interactions collected by Allen AI by offering free access to GPT-3.5 and GPT-4. It complements LMSYS as an independent sample of real-world prompts and ships with per-conversation metadata for language and toxicity, which we use during filtering. We draw 93 English prompts, including two independent real-world conversation sources rather than one, to guard against either corpus's idiosyncrasies skewing the conclusions.

**MMLU**, the Massive Multitask Language Understanding benchmark (Hendrycks et al., 2021), is a multiple-choice factual-knowledge benchmark spanning 57 subjects from elementary mathematics to professional law. Each item is a question with four answer options. We sample 100 items and treat them as a single factual-QA category, since the per-subject sample sizes would otherwise be too small for stable per-category statistics. MMLU represents the dense, fact-laden prompt at the opposite end of the spectrum from conversational chat.

**GSM8K** (Cobbe et al., 2021) is a benchmark of grade-school math word problems written in natural language, designed to test multi-step arithmetic reasoning. We sample 100 problems, all of which fall into a single math-reasoning category. These prompts are interesting for compression because they mix natural-language narrative (compressible) with numerically critical content (where dropping a word could destroy the problem), making them a good stress test for the safety mechanisms.

**HellaSwag** (Zellers et al., 2019) is a commonsense natural-language-inference benchmark in which the model must choose the most plausible continuation of a short context from four candidates. We sample 100 items, treated as a single commonsense category. As the results section shows, HellaSwag proves to be the most compression-sensitive of all our sources, which makes it valuable precisely as a failure case that delineates the limits of lexical compression.

## 4.2 Task categorisation

Dolly ships with task labels; the other sources do not. We assign each LMSYS, WildChat, and (initially) MMLU prompt to one of thirteen task categories using a rule-based classifier built on regex patterns calibrated against sample inspection. The categories are: code, math_reasoning, translation, summarization, creative_writing, information_extraction, classification, brainstorming, instructional, closed_qa, open_qa, general_qa, roleplay, safety_probe, commonsense, factual_qa, and a residual 'general' bucket. MMLU 57 subject labels are collapsed to a single factual_qa category to prevent fragmentation at small sample sizes. This taxonomy is heuristic; we treat per-category numbers as descriptive rather than statistically certified.

## 4.3 Configurations and metrics

Each prompt is passed through all fifteen toggle configurations. For each configuration, we call gpt-4o-mini twice -- once with the original prompt, once with the compressed prompt -- at temperature = 0. We then compute, for each (prompt, configuration) cell:

- Compression metrics: token reduction percent (cl100k_base) and character reduction percent.
- Prompt-side similarity: cosine similarity between original and compressed prompt under MiniLM-L6-v2.
- Output-side similarity (original-output vs compressed-output): BLEU-4 with smoothing (Papineni et al., 2002), ROUGE-1/2/L F1 (Lin, 2004), MiniLM cosine, and BERTScore-F1 (Zhang et al., 2020).

Caching is used to avoid redundant calls: each unique (prompt, configuration) pair is called exactly once; the original-prompt output is cached across all fifteen configurations for the same prompt. This reduces total OpenAI calls per prompt from 30 to approximately 16 (one original-output call plus one call per non-no-op compressed configuration).

### *4.3.1 What each metric measures*

Because the six output-side metrics each capture a different notion of similarity, we describe what each one means and how to read it. In every case the metric compares the model's response to the original prompt against its response to the compressed prompt; a high value means compression did not change what the model said.

Token reduction percent is the headline efficiency metric: the percentage decrease in cl100k_base token count from original to compressed prompt. This is the quantity a practitioner is trying to maximize, since input tokens are what API providers bill. Character reduction percent is the analogous figure on raw characters and is reported for completeness.

**Cosine similarity**, computed over MiniLM-L6-v2 SentenceBERT embeddings (Reimers & Gurevych, 2019), answers the question of whether the two outputs mean the same thing. It embeds each full output as a single vector and measures the cosine of the angle between them. It is the most forgiving of our metrics because it tolerates paraphrase: two outputs that convey identical meaning in entirely different words still score highly. Values above roughly 0.95 indicate essentially identical meaning; values below 0.70 indicate the outputs have genuinely diverged.

**BLEU-4** with smoothing (Papineni et al., 2002) measures n-gram precision, that is, whether the two outputs use the same words in the same order. Originally designed for machine translation, it is the strictest metric in our set: it penalizes any rewording even when meaning is preserved perfectly. A low BLEU score therefore does not necessarily indicate a quality problem; it often simply reflects that the language model rephrased the same answer. We include it for comparability with prior compression and generation literature.

**ROUGE-1, ROUGE-2, and ROUGE-L F1** (Lin, 2004) form a family of overlap metrics. ROUGE-1 measures unigram (single-word) overlap and so captures shared vocabulary regardless of order. ROUGE-2 measures bigram (consecutive word-pair) overlap and is correspondingly stricter, since it requires local word order to match. ROUGE-L measures the longest common subsequence, rewarding outputs that share a long ordered backbone of words even with other words interspersed, and thus captures structural similarity better than BLEU. Reading the three together gives a graded picture: ROUGE-1 high but ROUGE-2 low, for example, indicates the same vocabulary reorganized into different phrasing.

**BERTScore F1** (Zhang et al., 2020) is our headline quality metric. Rather than matching surface tokens, it embeds each token in a contextual BERT space and greedily matches each token in one output to its most similar token in the other, then reports the F1 of those soft matches. Because it operates on contextual embeddings rather than exact strings, it correlates more closely with human judgments of equivalence than BLEU or ROUGE, while being less coarse than whole-sentence cosine. Values above 0.95 indicate essentially equivalent outputs; 0.90 to 0.95 indicates very close outputs with minor phrasing differences; below 0.85 indicates meaningful divergence. When we summarise a configuration's quality in a single number, it is this one.

### 4.4 Persistence and resumability

Results are persisted to SQLite with a UNIQUE(prompt_id, config_id, run_id) constraint. The batch evaluator is resumable: interrupted runs are completed without redundant API spend. All compressed prompts, both GPT outputs, and per-stage compression traces are stored for post-hoc qualitative inspection. The total of 18,630 cells reported here completed with zero error rows.

## 5. Results

### 5.1 Headline configuration ranking

Table 1 summarises mean compression and similarity metrics for each of the fifteen configurations, ranked by a composite score equal to (mean token reduction) x (mean BERTScore-F1). Each row aggregates 1,242 prompts. All standard deviations were under sigma = 0.20 for cosine and under sigma = 0.05 for BERTScore-F1 except where noted in the discussion.

| # | Configuration | Tok red. (%) | Cos prompt | Cos out | BLEU | ROUGE-L F1 | BERT-F1 | Score |
|---|---|---|---|---|---|---|---|---|
| 1 | maximum | 40.25 | 0.741 | 0.728 | 0.157 | 0.281 | 0.876 | 0.352 |
| 2 | only_caveman | 35.19 | 0.907 | 0.847 | 0.308 | 0.445 | 0.910 | 0.320 |
| 3 | aggressive | 30.05 | 0.908 | 0.841 | 0.285 | 0.431 | 0.907 | 0.273 |
| 4 | only_stopwords | 29.62 | 0.933 | 0.857 | 0.319 | 0.460 | 0.913 | 0.270 |
| 5 | medium | 29.34 | 0.933 | 0.857 | 0.319 | 0.462 | 0.913 | 0.268 |
| 6 | only_function_words | 12.23 | 0.979 | 0.907 | 0.478 | 0.598 | 0.937 | 0.115 |
| 7 | light | 2.22 | 0.990 | 0.949 | 0.687 | 0.765 | 0.963 | 0.021 |
| 8 | only_filler_words | 2.19 | 0.990 | 0.950 | 0.694 | 0.772 | 0.964 | 0.021 |
| 9 | only_filler_phrases | 0.96 | 0.999 | 0.971 | 0.799 | 0.858 | 0.978 | 0.009 |
| 10 | normalize_only | 0.90 | 0.999 | 0.972 | 0.801 | 0.860 | 0.978 | 0.009 |
| 11 | baseline_none | 0.90 | 0.999 | 0.972 | 0.802 | 0.859 | 0.978 | 0.009 |
| 12 | only_contractions | 0.89 | 0.998 | 0.966 | 0.776 | 0.837 | 0.975 | 0.009 |
| 13 | only_abbreviations | 0.88 | 0.998 | 0.971 | 0.796 | 0.855 | 0.977 | 0.009 |
| 14 | only_lemmatize | 0.52 | 0.974 | 0.921 | 0.509 | 0.640 | 0.941 | 0.005 |
| 15 | only_synonyms | 0.23 | 0.841 | 0.808 | 0.316 | 0.445 | 0.909 | 0.002 |

*Table 1. Mean values per configuration across all 1,242 prompts. Score = (tok red. / 100) x BERT-F1.*

### *5.1.1 Interpretation*

The maximum configuration achieves the highest composite score, with 40.3% mean token reduction at 0.876 BERTScore-F1. However, its output cosine drops to 0.728, a noticeable degradation that we attribute to the inclusion of synonym shortening -- see section 5.3. The runner-up, only_caveman, achieves 35.2% reduction with output cosine 0.847 and BERTScore-F1 0.910, suggesting that POS-based pruning alone is the single most effective lever for aggressive compression. Configurations 4 and 5, only_stopwords and medium, are statistical twins: they achieve essentially identical compression (29.6% vs 29.3%) and identical output preservation (cosine 0.857, BERT 0.913), indicating that once stopwords are removed the additional medium-level transforms (abbreviations, contractions, filler words) contribute almost no further compression. This is consistent with the negligible scores of configurations 9 through 13.

## 5.2 Per-category Pareto behavior

Table 2 reports the medium configuration performance broken down by task category, restricted to categories with at least 50 result rows so that means are stable. The medium configuration is chosen for this breakdown because it sits at the upper-right of the headline Pareto frontier.

| Category | Prompts | Tok red. (%) | Cos out | BERT-F1 |
|---|---|---|---|---|
| open_qa | 100 | 32.5 | 0.916 | 0.929 |
| code | 99 | 26.3 | 0.884 | 0.928 |
| general_qa | 100 | 34.5 | 0.903 | 0.922 |
| summarization | 100 | 27.7 | 0.911 | 0.921 |
| classification | 100 | 17.2 | 0.909 | 0.918 |
| information_extraction | 102 | 25.6 | 0.846 | 0.916 |
| math_reasoning | 117 | 30.2 | 0.909 | 0.915 |
| closed_qa | 100 | 28.0 | 0.863 | 0.914 |
| factual_qa | 100 | 28.9 | 0.863 | 0.907 |
| creative_writing | 100 | 36.0 | 0.834 | 0.904 |
| commonsense | 100 | 34.8 | 0.674 | 0.880 |

*Table 2. The medium configuration evaluated per task category, ordered by BERT-F1 (descending).*

### *5.2.1 Category-by-category analysis*

We discuss each category in descending order of preserved quality (the order of Table 2). Throughout, recall that BERTScore-F1 is our headline quality figure and cosine is a complementary whole-output meaning check.

**Open QA** (BERT-F1 0.929, cosine 0.916, 32.5% reduction) is the best-preserved category. These are open-ended questions whose answers, while free-form, converge on a stable core of facts. Even after a third of the prompt tokens are removed, the model recovers the same answer, because the essential question survives in the content words. Open QA is close to an ideal case for lexical compression: high reduction at near-top fidelity.

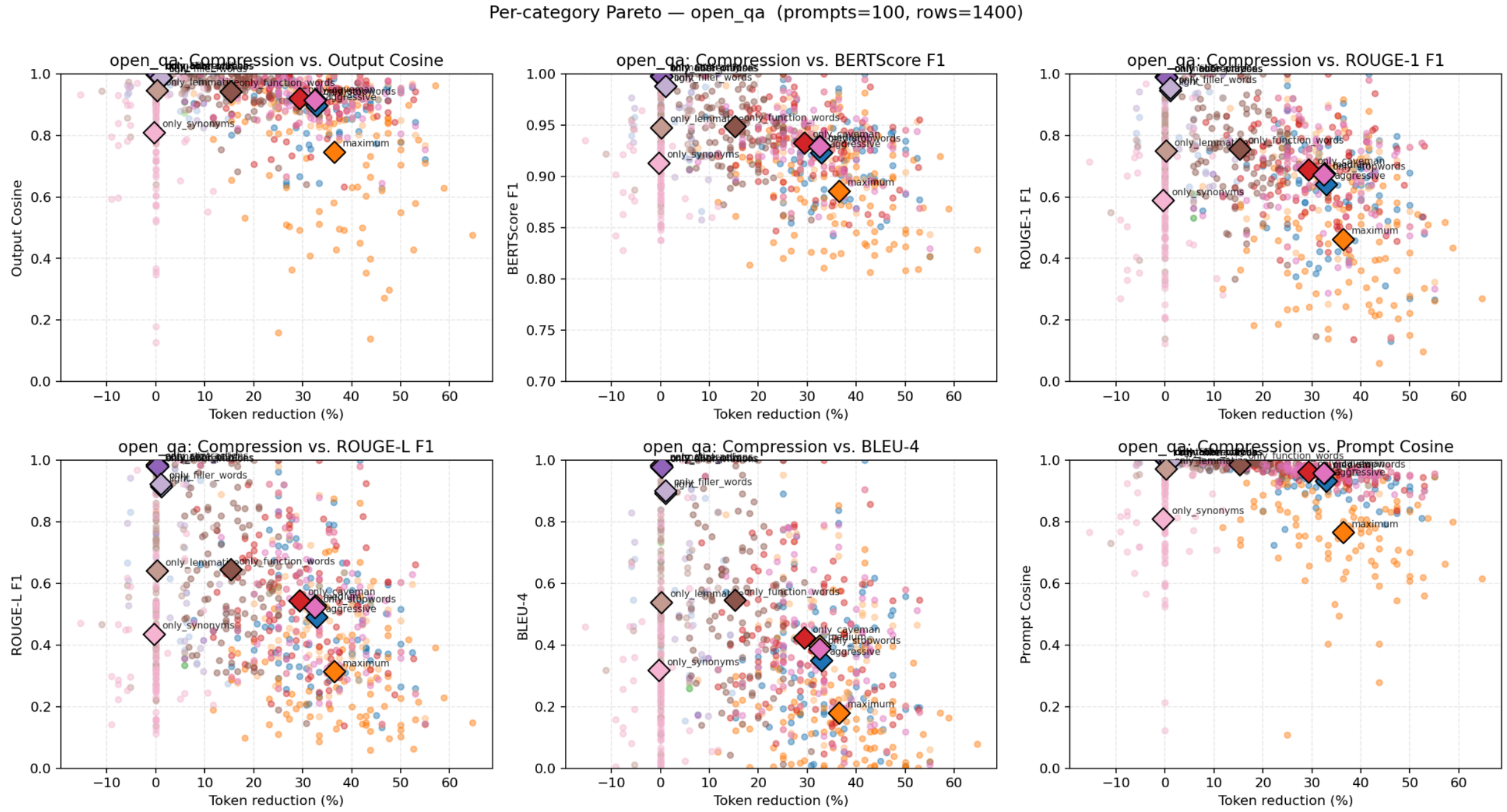

**Code** (BERT-F1 0.928, cosine 0.884, 26.3% reduction) preserves quality very well. Code-related prompts carry their intent in identifiers, keywords, and technical nouns, which the content-preserving techniques retain. The slightly lower cosine than open QA reflects that generated code can vary in formatting and comments while remaining functionally equivalent, which whole-output cosine penalizes more than BERTScore does.

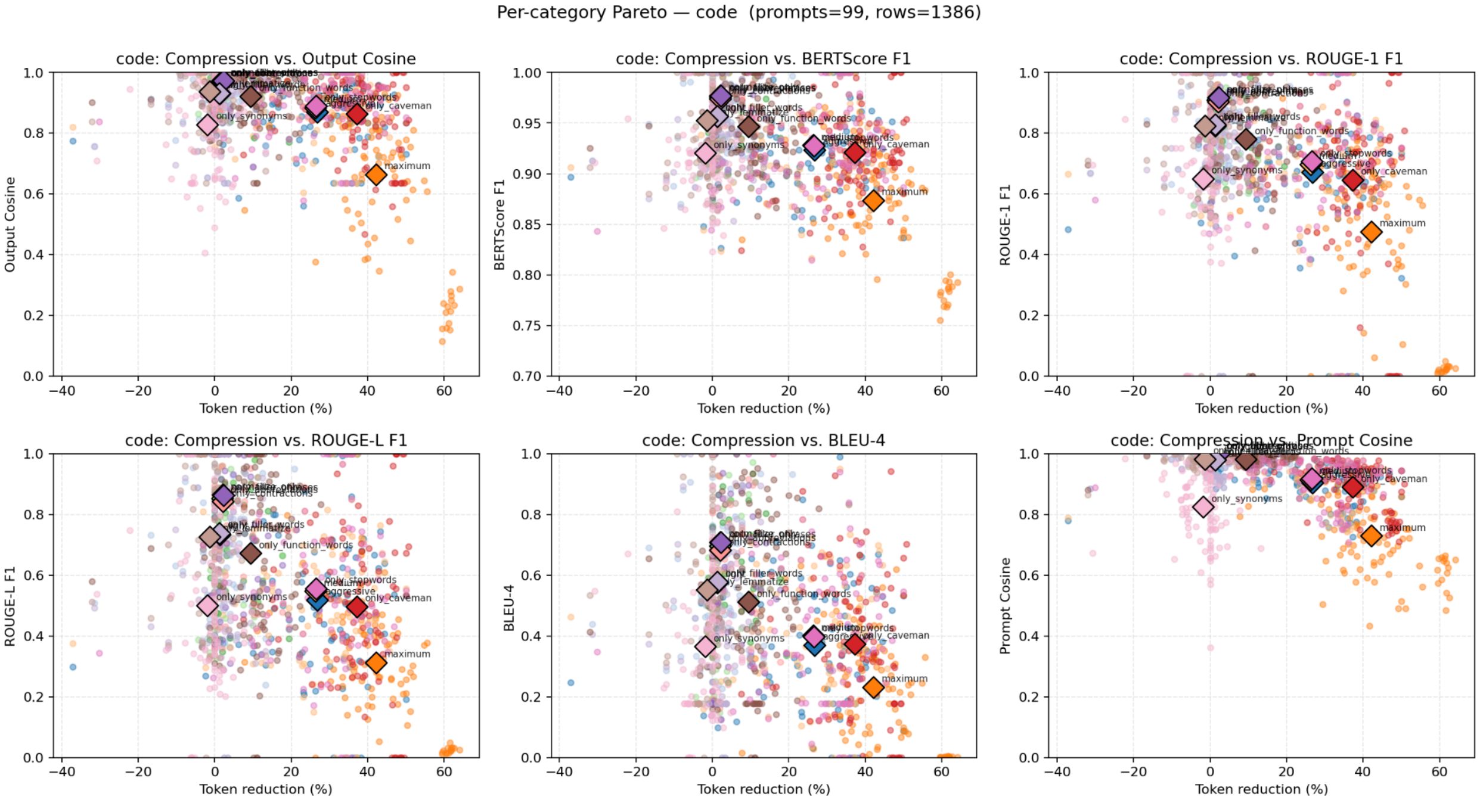

**General QA** (BERT-F1 0.922, cosine 0.903, 34.5% reduction) combines the second-highest compression with strong preservation. Like open QA, the answer space is constrained enough that a leaner prompt elicits the same response. The high reduction reflects that general questions are often phrased verbosely, leaving ample filler to remove.

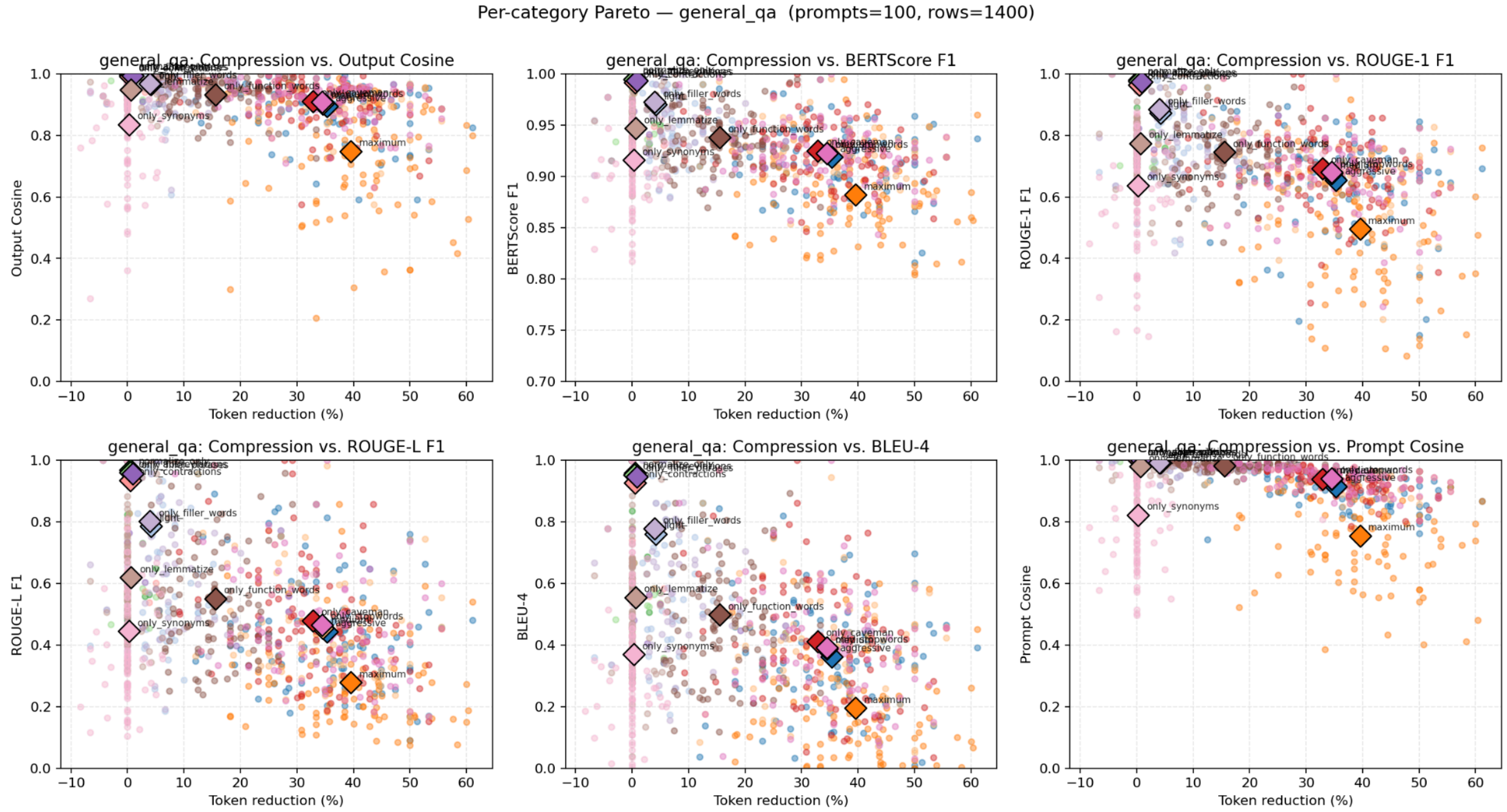

**Summarization** (BERT-F1 0.921, cosine 0.911, 27.7% reduction) holds up well. The instruction to summarise plus the source content compresses cleanly, and because the task itself is to distill meaning, the model is robust to a terser prompt. Cosine is high here, indicating the summaries themselves remain close in meaning.

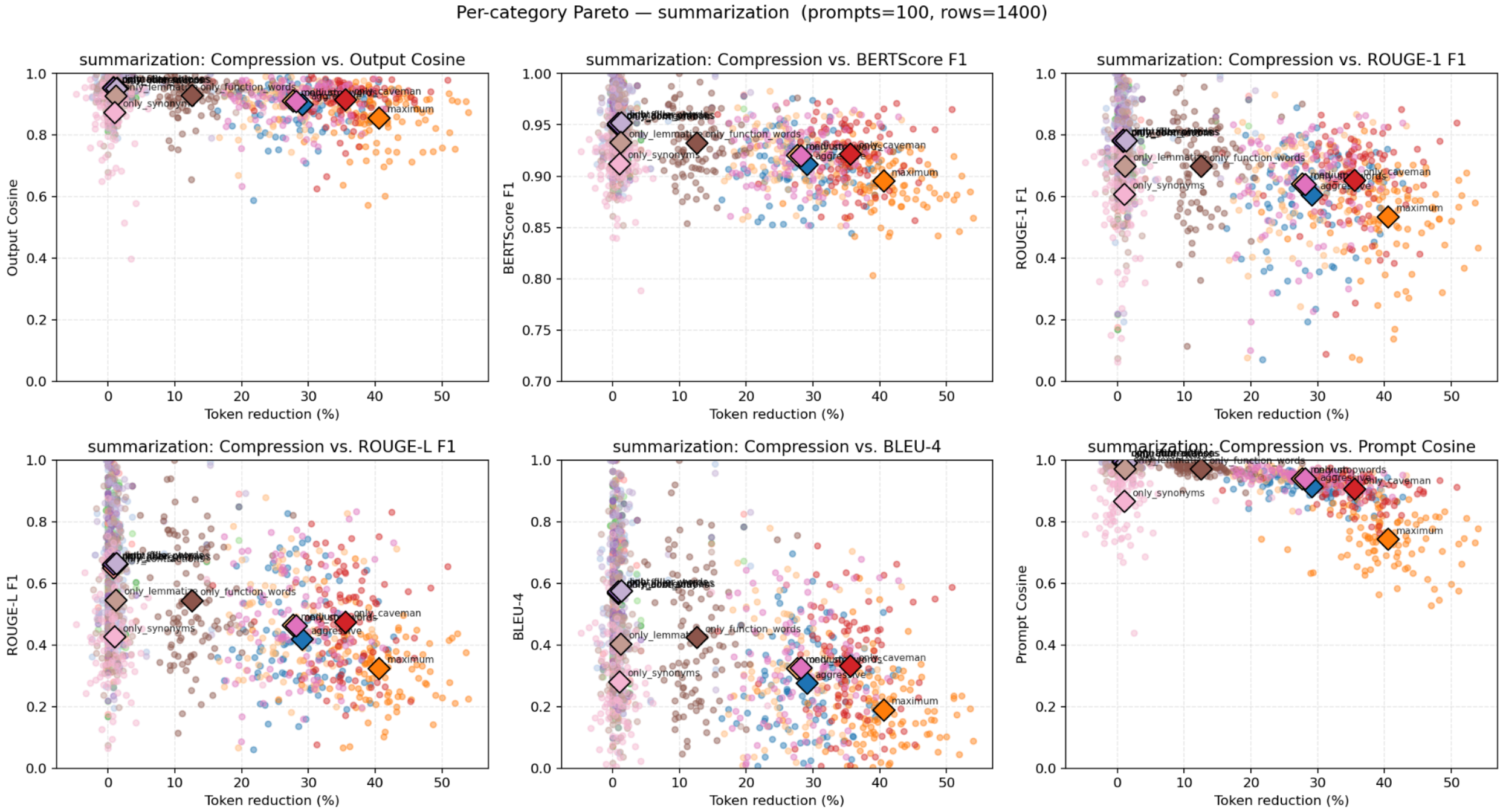

**Classification** (BERT-F1 0.918, cosine 0.909) shows the lowest reduction of any category at 17.2%. This is expected: classification prompts are often already terse (a short instruction plus the item to classify) and carry little removable filler. The compression that does occur is safe, so quality stays high. Classification illustrates that low reduction is not a failure; it simply reflects a prompt with little fat to trim.

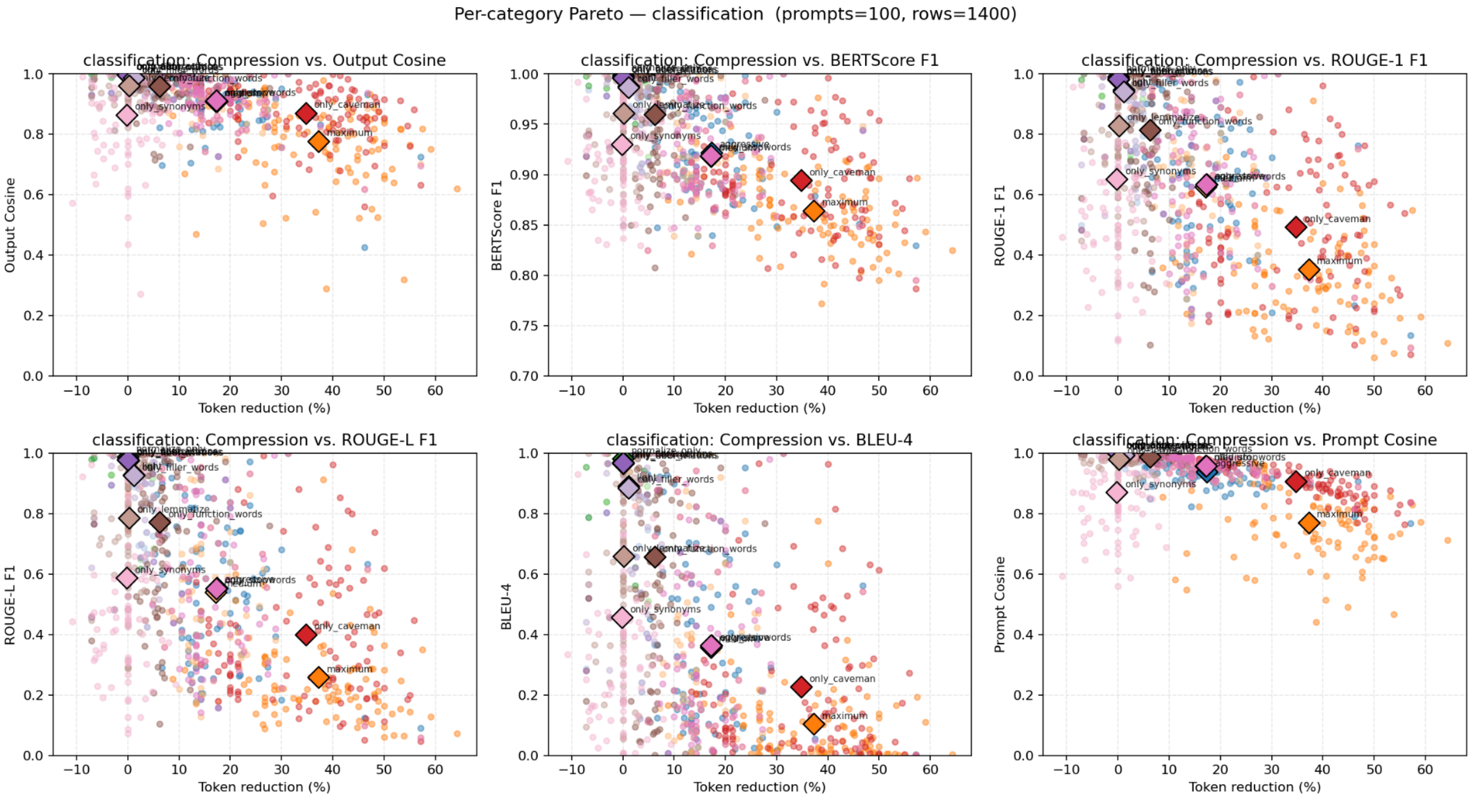

**Information extraction** (BERT-F1 0.916, cosine 0.846, 25.6% reduction) preserves the BERTScore quality band but shows a noticeably lower cosine. The lower cosine arises because extraction outputs are short and structured, so any difference between the two outputs moves the whole-output vector substantially, even when the extracted facts are equivalent. BERTScore, matching token by token, correctly reports that the content is preserved.

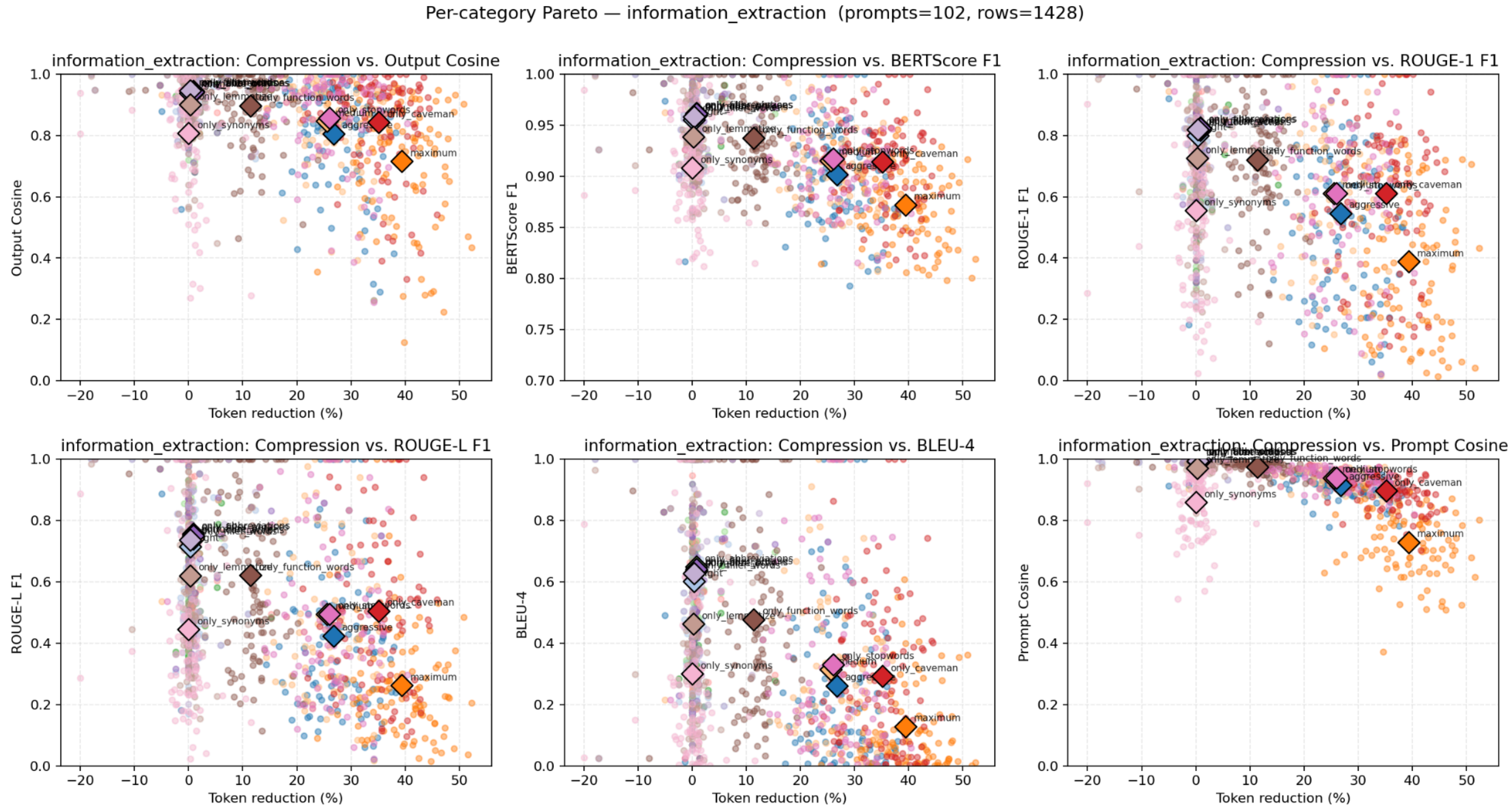

**Math reasoning** (BERT-F1 0.915, cosine 0.909, 30.2% reduction) is more robust than one might expect for a reasoning task. The GSM8K narrative wrapping around the numbers is compressible, while the safety mechanisms keep the numerically critical tokens and negations intact. The model reaches the same answer from the leaner prompt in the large majority of cases, though this is the category where the occasional dropped connective word does the most damage, which is reflected in its position toward the lower-middle of the table.

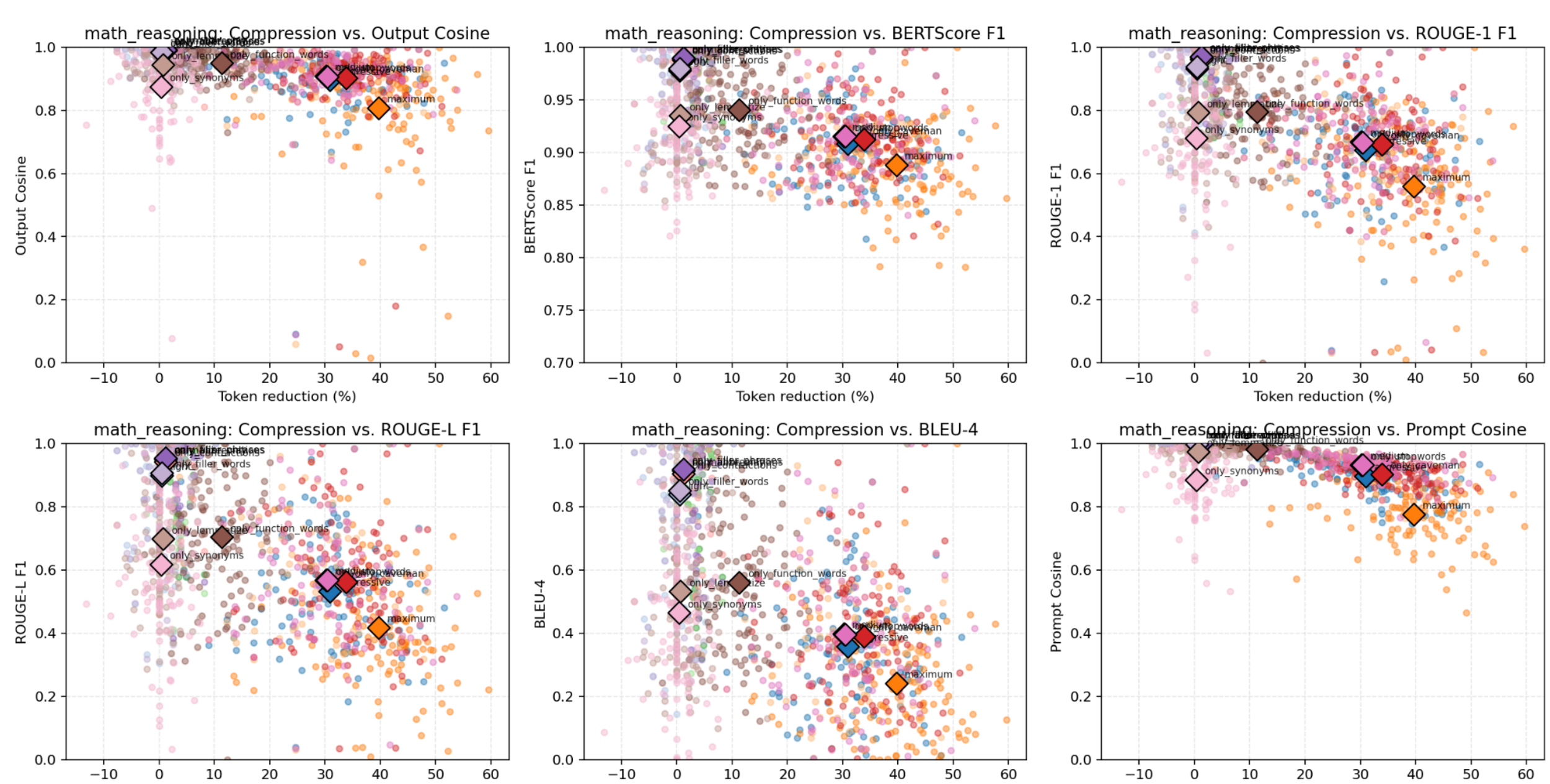

**Closed QA** (BERT-F1 0.914, cosine 0.863, 28.0% reduction) and factual QA (BERT-F1 0.907, cosine 0.863, 28.9% reduction, the latter being our consolidated MMLU category) sit together in the lower-middle band. Both involve retrieving a specific fact, and both show a gap between solid BERTScore and more modest cosine. The pattern is the same as information extraction: short factual answers make cosine volatile while the underlying facts, and thus BERTScore, remain stable. These categories tolerate roughly a 28% reduction without meaningful quality loss.

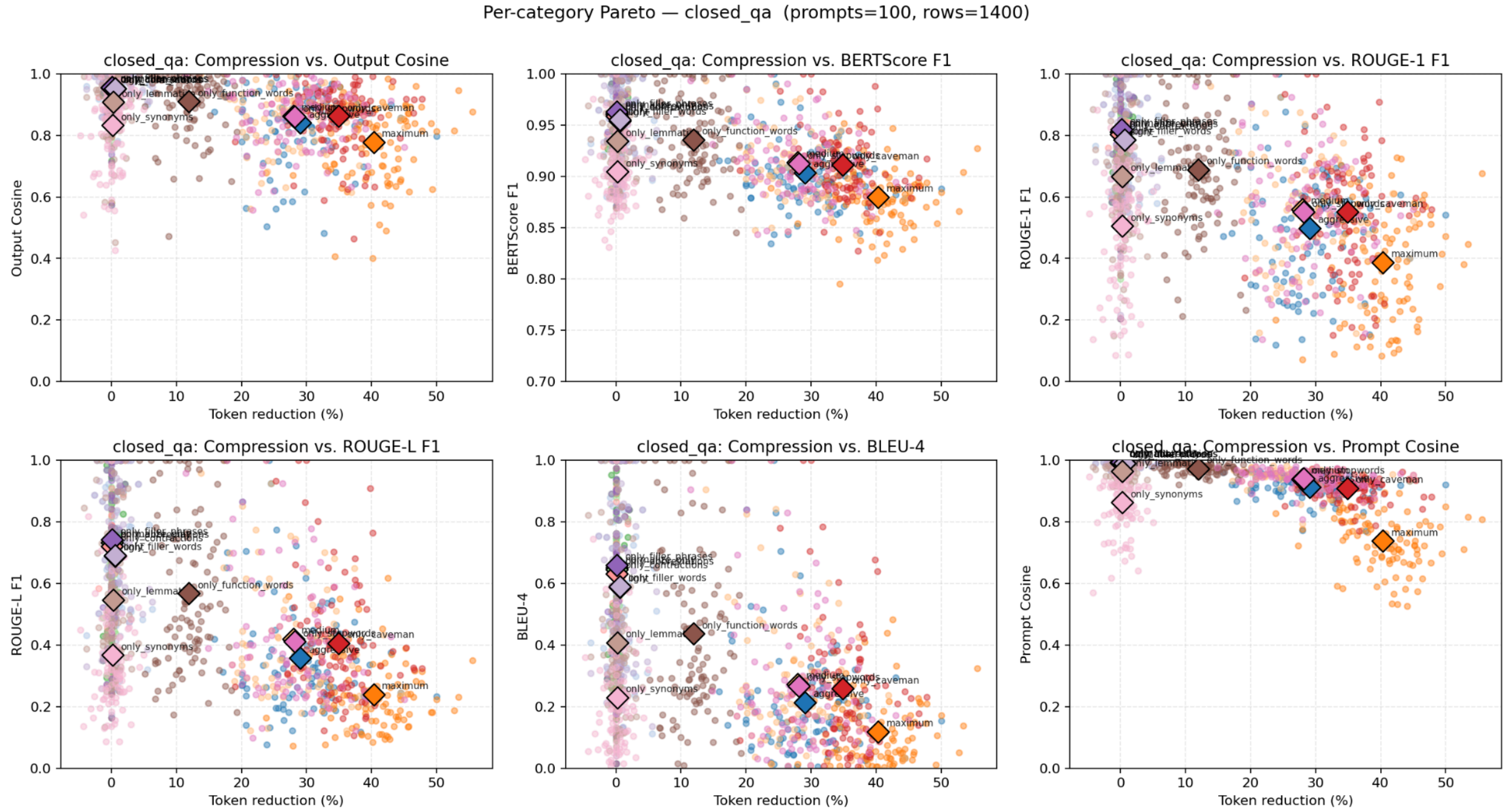

**Creative writing** (BERT-F1 0.904, cosine 0.834, 36.0% reduction) achieves the highest compression of any category but the second-lowest cosine. This is not a failure mode but an intrinsic property of the task. A creative brief admits an enormous space of valid outputs, so any perturbation of the prompt, including compression, nudges the model to a different but equally valid story. BERTScore remains respectable because the outputs share theme and vocabulary; cosine drops because the two stories are structurally distinct. For creative tasks, the user should expect a different-but-valid result, and high reduction is safe on that understanding.

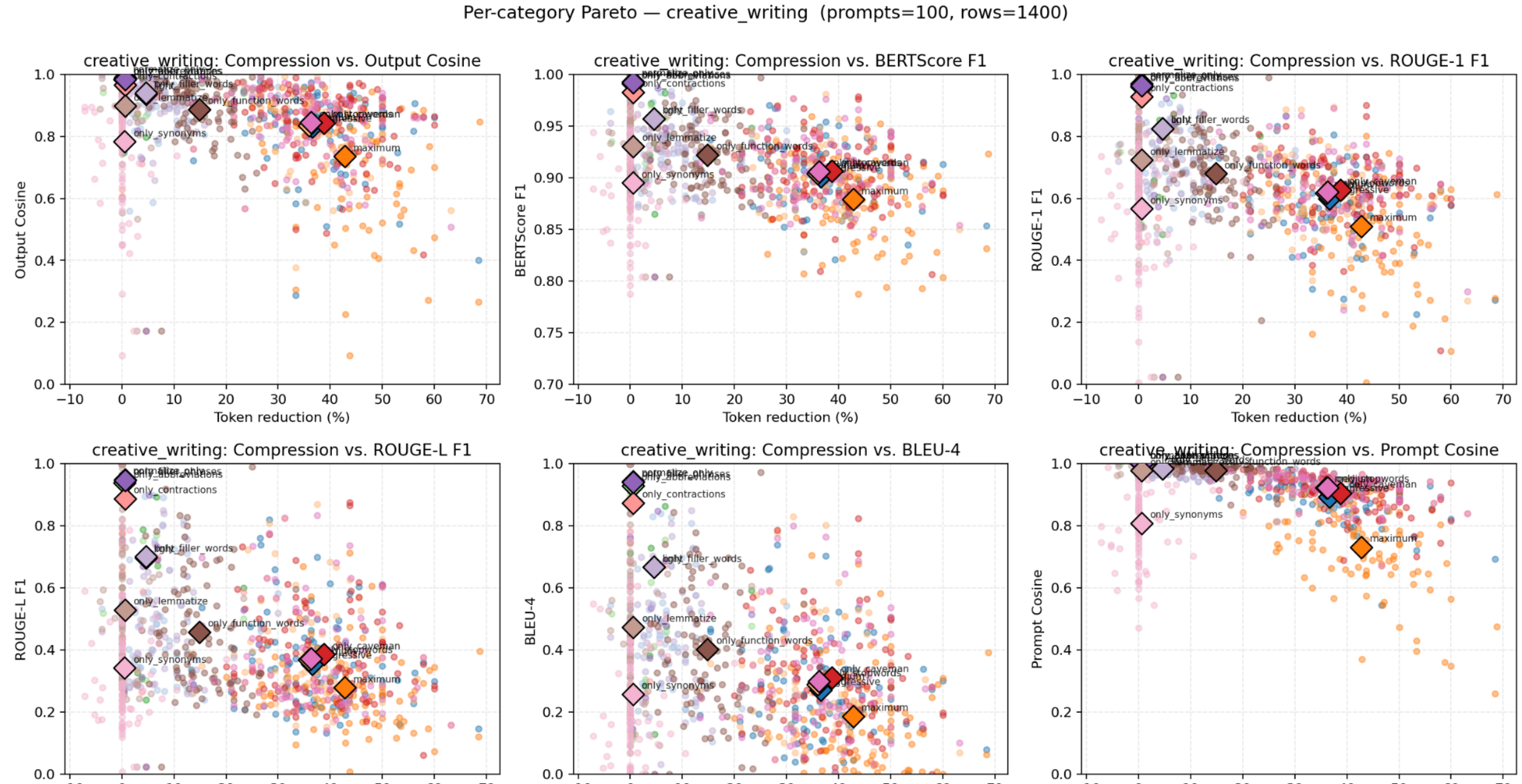

**Commonsense** (BERT-F1 0.880, cosine 0.674, 34.8% reduction), drawn from HellaSwag, is the clear failure mode and the most informative result in the study. Its cosine of 0.674 is far below every other category. HellaSwag items require choosing the most plausible sentence continuation, a judgment that depends on a dense web of common nouns, articles, and short function words that establish the precise situation. Aggressive removal of stopwords and function words strips exactly the disambiguating cues the task depends on, so the model often selects a different continuation. This is consistent with the broader observation that reasoning tasks are sensitive to token-level perturbation (Cobbe et al., 2021). The practical lesson is that commonsense-inference prompts should be compressed conservatively if at all.

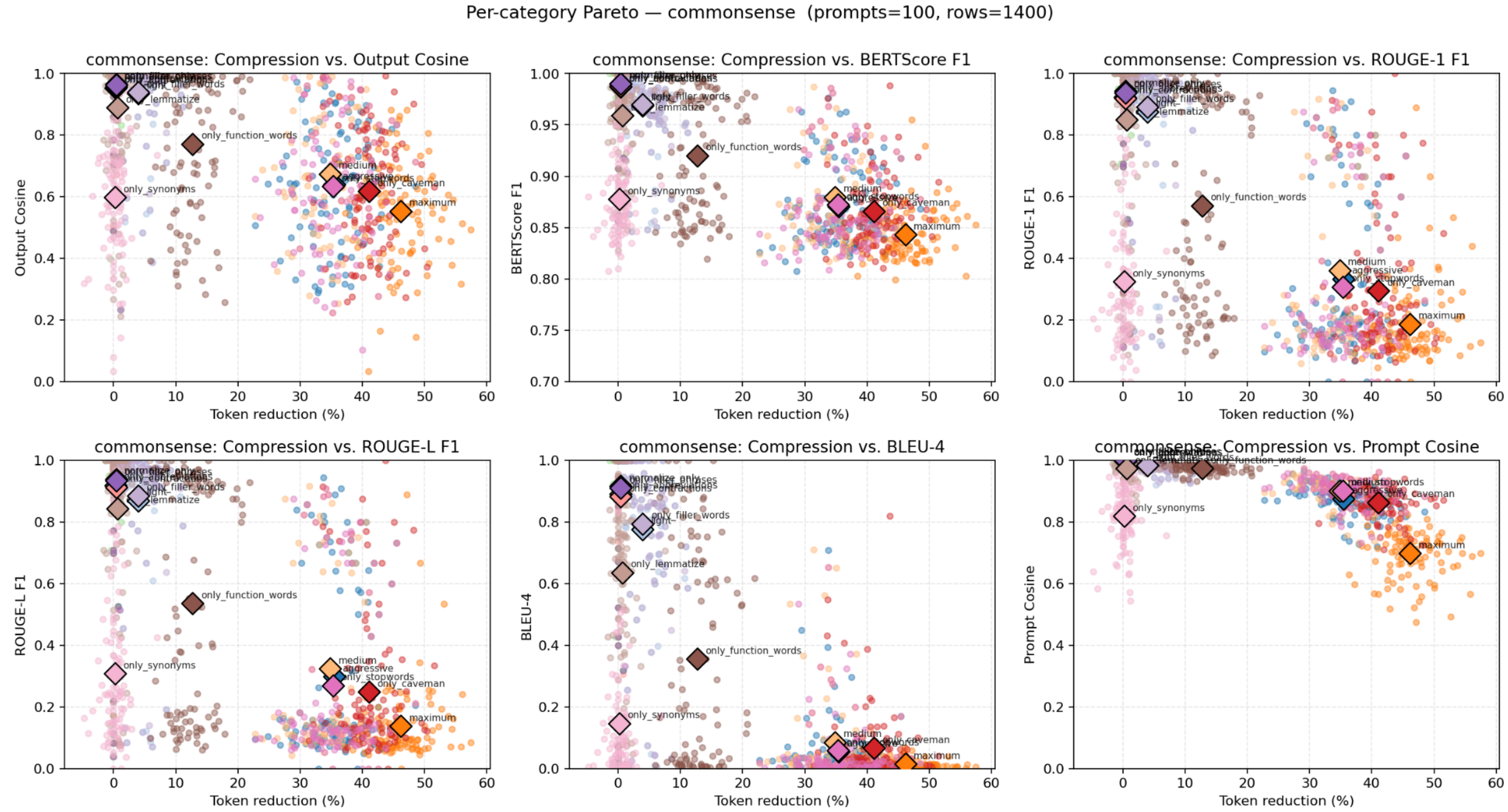

### *5.2.2 Summary of the per-category pattern*

Two regularities emerge. First, categories whose answers occupy a narrow space (open QA, general QA, code, summarization, math reasoning) tolerate aggressive compression because the model reconverges on the same answer. Second, the gap between BERTScore and cosine widens for categories with short, structured outputs (information extraction, closed and factual QA), where cosine is volatile but BERTScore confirms content is preserved. The genuine outlier is commonsense inference, where the task's reliance on dense low-level cues makes it the one category lexical compression should approach with caution.

## 5.3 The synonym-shortening pathology

The only_synonyms configuration (row 15 in Table 1) is anomalous: it achieves only 0.23% mean token reduction yet drives prompt cosine down to 0.841 and output BERTScore-F1 down to 0.909 -- the lowest among single-technique configurations. Inspection reveals that WordNet nearest-by-token-length synonym is frequently a topically related but semantically narrower term (e.g., 'house' to 'family', 'large' to 'big' in many contexts but 'big' to 'great' in others). The cl100k_base tokenizer also sometimes assigns equal token counts to forms that differ in usage frequency, leaving the substitution decision underconstrained. The technique should not be used in isolation; we retain it in the configuration sweep so that its weakness is empirically documented.

## 5.4 Recommended operating points

From RQ4: subject to a BERT-F1 floor of 0.90 (a generally accepted threshold for 'near-paraphrase' output equivalence), the most aggressive configuration is only_caveman (35.2% reduction, BERT-F1 0.910). For practitioners unwilling to use the heavier POS-keep transform, medium or only_stopwords (each 29.6% reduction, BERT-F1 0.913) are very nearly equivalent and operationally simpler to defend. We recommend these as the two practical operating points for downstream use.

# 6. Discussion

## 6.1 What the Pareto curve looks like

Plotting all 1,242 prompts at all fifteen configurations against the BERTScore-F1 axis yields a clear monotone Pareto front: the configurations align along a smooth concave envelope from approximately 0% reduction at BERT-F1 around 0.98 down to approximately 40% reduction at BERT-F1 around 0.87. There is no configuration that achieves both >35% reduction and >0.91 BERT-F1; this appears to be an empirical ceiling for purely lexical, training-free compression on prompts of the lengths we tested. Future work that wishes to push beyond this point will need either token-level informativeness scoring (per LLMLingua) or learned compression.

## 6.2 Comparison to learned methods

We do not directly outperform LLMLingua, which reports up to 20x compression on long-context benchmarks. Our results are nevertheless informative because they hold the same metric framework constant while removing the auxiliary-model component. Under temperature = 0 and gpt-4o-mini, our maximum configuration (40.3% reduction, BERT-F1 0.876) compares favorably with the operating points LLMLingua reports for short-prompt scenarios. Because our pipeline is deterministic and CPU-only, it can be deployed as a zero-cost, audit-friendly first-stage filter before a heavier learned compressor.

## 6.3 Limitations

- The evaluator is one LLM. All calls to eval are made against gpt-4o-mini. Different evaluators (e.g., Claude, Gemini, a larger GPT model) may produce different BERTScore-F1 numbers. The directionality of the Pareto curve should be stable, but the absolute numbers are specific to the evaluator.
- I will be brief and medium prompt. We filter prompts longer than 2000 characters. The dynamics for extremely long retrieval-augmented prompts are not characterized here, this is the regime where learned compression (LongLLMLingua) has reported its strongest gains.
- English only. The langdetect-based filter and the WordNet lexicon are both English-centric. Multilingual extension is open work.

- Category labels are heuristic. Per-category numbers should be interpreted as descriptive. Manual relabelling of a subset and inter-annotator agreement assessment are deferred to future work.
- Compression-only latency is not reported. The 18,630 cells were recorded with end-to-end timings that include two OpenAI API calls each. We did not separately benchmark the compression component itself; a dedicated latency study is reserved for the public release of the toolkit.

### 6.4 Threats to validity

Two threats worth flagging. First, our automatic categorizer uses keyword regexes that we tuned on a small inspection set, and categories with few keyword cues (general, instructional) may have misclassifications. Second, ROUGE and BLEU are sensitive to surface-form changes – a compressed prompt producing a correctly paraphrased answer will score lower on these metrics than on BERTScore-F1, which is why BERTScore-F1 (and to a lesser extent cosine) is the headline metric we report.

## 7. Conclusion

We have proposed a deterministic, training-free, CPU-only lexical compression pipeline for LLM prompts, characterized its compression-vs-fidelity Pareto frontier on 18,630 paired GPT-4o-mini completions across sixteen task categories, and identified the configurations occupying the practically useful upper-right corner of that frontier. Our main empirical result is that lexicon-based methods can reduce the tokens by about 30% and keep 0.91 BERTScore-F1 against the output of the original prompt, without any auxiliary model and without any GPU. At a measurable but not catastrophic quality cost, it's possible to go further into the 35-40% range. The main residual failure mode is commonsense reasoning (HellaSwag), in line with previous results that token-level perturbations harm tasks with dense disambiguation cues.

Our results do not replace learned methods, which still have a clear quality advantage at very high compression ratios and on long context tasks. They do however provide a well-defined, reproducible baseline for the evaluation of future learned methods, and a practical solution for engineers who want to perform lossless-enough prompt compression today, on a laptop CPU, with no neural component in the loop.

## 8. Reproducibility

The full toolkit, including the compressor, a FastAPI service, a Streamlit UI, a batch evaluator, a categorization pipeline, a plotting suite and a SQLite results store, will be published as an open-source repository upon paper acceptance. All experiments are run with a fixed random seed (42). All API calls are at temperature 0. We include the SQLite database from our reported run so that all figures and tables can be regenerated end-to-end with a single command.

Dependencies pinned in requirements.The pipeline runs on Linux, macOS and Windows using Python 3.10 or later.

### 8.1 Public Python package

Beyond the research toolkit, the core compression pipeline is released as a standalone, installable Python package named less-tokens, available on the Python Package Index and installable with a single command (pip install less-tokens). The package is intentionally lightweight and dependency-minimal so that it can be dropped into production systems, and it is the recommended way to reproduce or build on the compression results reported here. It is released under the permissive MIT license.

The package exposes the same eleven techniques described in Section 3 through a single compress function, in which each technique is an independent on/off flag, so any of the fifteen configurations evaluated in this paper can be reconstructed exactly by setting the corresponding flags. The deterministic, training-free design guarantees that a given prompt-and-flag combination always yields byte-identical output, which makes the package suitable for both reproducing our cells and caching compressed prompts in deployment.

Two further entry points extend the pipeline for practical use. A compare function computes all six output-similarity metrics used in this paper (cosine, BLEU-4, ROUGE-1, ROUGE-2, ROUGE-L, and BERTScore-F1) given an original and a compressed prompt, along with the two corresponding model outputs, allowing users to replicate our evaluation protocol on their own prompts and their choice of language model. A compress_structured function addresses the practical reality that many real prompts contain segments that must not be compressed: it accepts a prompt divided into zones, each tagged as free (fully compressed), careful (compressed only with the meaning-preserving subset of techniques), or protected (returned verbatim). This lets a practitioner aggressively compress an instruction body while leaving an output schema, a set of rules, or worked examples untouched. Asynchronous variants of both compression functions are provided for high-throughput and event-loop-based deployments, and a command-line interface exposes the same capabilities for shell-based and scripting workflows.

Because the package and the research toolkit share the same underlying compression code, results obtained with the published package are directly comparable to the numbers reported in this paper. We therefore recommend the package as the canonical reference implementation for both reproduction and downstream application.